\documentclass[runningheads]{llncs}
\usepackage[T1]{fontenc}
\usepackage{graphicx}
\usepackage{multirow} 
\usepackage{amsmath}
\begin{document}
\title{Large Language Models as Universal Predictors? An Empirical Study on Small Tabular Datasets}
\titlerunning{LLMs as Universal Predictors for Tabular Data}
%
\author{Nikolaos Pavlidis\inst{1,2}\orcidID{0000-0001-9370-5023} \and
Vasilis Perifanis\inst{1,2}\orcidID{0000-0003-3915-9628}
 \and
Symeon Symeonidis\inst{3}\orcidID{0000-0002-3259-614X}
\and
Pavlos S. Efraimidis\inst{1,2}\orcidID{0000-0003-3749-0165}
}
\authorrunning{N. Pavlidis et al.}
%
\institute{Institute for Language and Speech Processing, Athena Research Center, 67100 Xanthi, Greece \and
Department of Electrical and Computer Engineering, Democritus University of Thrace, Kimmeria, 67100 Xanthi, Greece \and 
Department of Production and Management Engineering, Democritus University of Thrace, Kimmeria, 67100 Xanthi,Greece
}
\maketitle              
\begin{abstract}
Large Language Models (LLMs), originally developed for natural language processing (NLP), have demonstrated the potential to generalize across modalities and domains. With their in-context learning (ICL) capabilities, LLMs can perform predictive tasks over structured inputs without explicit fine-tuning on downstream tasks. In this work, we investigate the empirical function approximation capability of LLMs on small-scale structured datasets for classification, regression and clustering tasks. We evaluate the performance of state-of-the-art LLMs (GPT-5, GPT-4o, GPT-o3, Gemini-2.5-Flash, DeepSeek-R1) under few-shot prompting and compare them against established machine learning (ML) baselines, including linear models, ensemble methods and tabular foundation models (TFMs). Our results show that LLMs achieve strong performance in classification tasks under limited data availability, establishing practical zero-training baselines. 
In contrast, the performance in regression with continuous-valued outputs is poor compared to ML models, likely because regression demands outputs in a large (often infinite) space, and clustering results are similarly limited, which we attribute to the absence of genuine ICL in this setting. Nonetheless, this approach enables rapid, low-overhead data exploration and offers a viable alternative to traditional ML pipelines in business intelligence and exploratory analytics contexts. We further analyze the influence of context size and prompt structure on approximation quality, identifying trade-offs that affect predictive performance. Our findings suggest that LLMs can serve as general-purpose predictive engines for structured data, with clear strengths in classification and significant limitations in regression and clustering.

\end{abstract}
\section{Introduction}
Large-scale pre-trained language models (LLMs) like GPT~\cite{achiam2023gpt} and Gemini~\cite{team2023gemini}, have achieved state-of-the-art results in a wide range of natural language processing (NLP) tasks, such as translation, summarization, question answering and code generation. These capabilities have raised interest in the broader potential of LLMs to act as general-purpose function approximators across domains beyond natural language.

From a theoretical perspective, the universal approximation theorem guarantees that sufficiently wide feed-forward neural networks with non-linear activations can approximate any function on compact domains~\cite{hornik1991approximation}. However, the architectures and training objectives of modern LLMs, especially for transformer models trained in an autoregressive fashion, deviate substantially from traditional settings. Hence, it remains an open question to what extent LLMs can approximate functions over structured, non-linguistic data such as tabular inputs, especially in scenarios where explicit model fine-tuning is not performed.

Recent developments in in-context learning (ICL)~\cite{brown2020language}\cite{shamshiri2024context} demonstrate that LLMs can solve a variety of unseen tasks by conditioning on a few examples within a prompt, without requiring fine-tuning processes. The fundamental concepts and working mechanism of ICL are outlined in ~\cite{dong-etal-2024-survey}. This ability, along with the increasing amount of tabular data in domains like healthcare~\cite{briola2024federated}, has given efforts to explore LLMs for tabular tasks. Although previous works have explored table serialization and fine-tuning for structured predictions ~\cite{lu2025fine}~\cite{jaitly2023towards}, fundamental limitations remain regarding input length constraints, numerical reasoning capabilities and the ability to generalize to diverse features~\cite{dong2024large}.

Demonstrating that LLMs can effectively approximate functions in structured data tasks presents the potential for developing a universal baseline predictor that operates without the need for explicit model training. For instance, by supplying a few representative examples for supervised tasks, LLMs could perform predictive and analytical tasks without traditional machine learning (ML) pipelines. This paradigm has practical applications in business intelligence as it would enable rapid exploratory data analysis, reduce the time required to derive insights, and provide non-technical users with advanced analytics capabilities without coding or data science expertise.


In this work, we empirically evaluate the approximation capability of LLMs on small-scale structured data tasks, without task-specific fine-tuning. We assess GPT-5 (low and high budget thinking) GPT-4o~\cite{achiam2023gpt}, GPT-o3, DeepSeek-R1~\cite{liu2024deepseek} and Gemini-Flash-2.5~\cite{team2023gemini}, using few-shot ICL on classification, regression and clustering tasks, leveraging datasets of up to 500 rows. Our goal is to assess whether LLMs can function as zero-training predictors, i.e., deriving predictive quality only from prompt conditioning, without explicit model optimization.

We further compare LLMs to traditional ML algorithms such as linear regression (LR), Random Forests (RF), LightGBM (LGBM), and more recent developments using tabular foundation models (TFMs) like TabPFN~\cite{hollmann2022tabpfn} and TabICL~\cite{qu2025tabicl}. Beyond accuracy metrics, we investigate the effect of varying the number and structure of context examples on LLM approximation quality.

The contributions of this work are summarized as follows:
\begin{itemize}
    \item We empirically evaluate LLMs (GPT-5, GPT-4o, GPT-o3, DeepSeek, Gemini) as black-box predictors on structured data tasks and compare them against well-established ML baselines.
    \item We analyze how the number of input examples and prompt context influence the ability of LLMs to generalize in classification settings.
    \item We demonstrate that LLMs can serve as practical, zero-training baseline predictors for classification tasks.
\end{itemize}

The rest of this paper is structured as follows. Section~\ref{sec:related_work} reviews related work on LLMs and TFMs for structured data. Section~\ref{sec:methodology} describes our experimental setup. Section~\ref{sec:experimental} presents the experimental results and analysis. Finally, Section~\ref{sec:conclusions} concludes this work and outlines future directions.

\section{Related Work}
\label{sec:related_work}
The emergence of large scale pre-trained models has transformed the landscape of deep learning~\cite{wen2024innovative}. These models, which are trained on diverse and massive datasets, have demonstrated the ability to perform a wide range of downstream tasks via ICL, a mechanism where the model can reason over input-output examples provided in the prompt, without any fine-tuning~\cite{shin2022effect}. 


The theoretical basis for treating deep models as general-purpose function approximators have its origin from the universal approximation theorem, which states that sufficiently wide neural networks with non-linear activations can approximate any continuous function on compact domains~\cite{hornik1991approximation}. However, these results were derived for fully connected architectures and do not directly generalize to the operational characteristics of modern transformer models.

Transformers, introduced by Vaswani et al.~\cite{vaswani2017attention}, rely on self-attention mechanisms rather than convolutional or recurrent structures, and are trained on sequence modeling objectives. Recent theoretical works have begun to explore the functional capacity of attention-based networks~\cite{kratsios2021universal}, but practical demonstrations of transformers as universal approximators, especially for structured and non-sequential data, are still limited.

Recent works have explored the application of LLMs to structured data, typically by converting tabular inputs into serialized textual formats. These approaches have enabled tasks such as code generation, table question answering and basic data transformations~\cite{lu2025large}. However, challenges remain in scenarios involving high-dimensional numeric features, numerical reasoning and generalization across diverse feature types or distributions~\cite{gorishniy2021revisiting}. 

To address the limitations of prompt serialization and LLMs’ context window size, the concept of tabular foundation models (TFMs) has recently gained attention. For example, TabPFN is a transformer model pretrained on synthetic datasets to perform classification tasks via ICL~\cite{hollmann2022tabpfn}. Its successor, TabPFNv2, improves both scalability and predictive performance~\cite{cheng2025realistic}. 


Although the recent advances in predictive modeling with deep architectures, traditional methods such as linear models, decision trees and gradient-boosting algorithms (e.g., LightGBM, CatBoost) continue to serve as competitive baselines for tabular tasks~\cite{anghel2018benchmarking}. Automated machine learning (AutoML) frameworks have further optimized performance in tabular scenarios by automatically tuning hyperparameters and selecting appropriate models~\cite{he2021automl}. Despite their strong predictive performance, these approaches still require explicit model training, feature engineering and computational resources.

The potential for LLMs to serve as zero-training predictors, leveraging ICL without any optimization, offers a compelling alternative to traditional pipelines. While prior works on tabular ICL, such as TabPFN, have focused on fine-tuned architectures, relatively little is known about the effectiveness of general-purpose LLMs in approximating functions over tabular data using only in-context examples. This study explores this potential by empirically evaluating the extent to which LLMs, prompted with structured examples, can compete with traditional ML baselines across classification, regression and clustering tasks.

\section{Methodology}
\label{sec:methodology}
This section outlines the datasets used for evaluation, the ML models considered and the evaluation metrics employed to assess predictive performance.

\subsection{Datasets}
We evaluate LLMs as general predictors on a curated set of small-scale tabular datasets commonly used in ML benchmarks. Each dataset is selected based on its limited sample size (maximum of 500 instances). The characteristics of the datasets considered for each task are presented in Table~\ref{tab:datasets}.

\begin{table}[t!]
\centering
\caption{Datasets Characteristics}
\label{tab:datasets}
\begin{tabular}{cccc}
\hline
\textbf{Dataset} & \textbf{\# Samples} & \textbf{\# Features} & \textbf{Task}  \\ \hline
Iris             & 150                 & 4                    & Classification \\
Lupus            & 87                  & 5                    & Classification \\
Bankrupt         & 250                 & 6                    & Classification \\
Diabetes         & 442                 & 10                   & Regression     \\
Servo            & 167                 & 4                    & Regression     \\
Friedman         & 500                 & 10                   & Regression     \\
Mall             & 200                 & 4                    & Clustering     \\
Wholesale        & 440                 & 7                    & Clustering     \\
Moon             & 300                 & 2                    & Clustering     \\ \hline
\end{tabular}
\end{table}

\paragraph{Classification}

\begin{itemize}
    \item \textbf{Iris}~\cite{fisher1936use}. A well-known three-class classification dataset with 150 samples and 4 features, used for predicting the species of iris flowers.
    \item \textbf{Lupus}~\cite{10.1093/bioinformatics/btab727}. A binary classification dataset with measurements from 87 patients, used to predict the presence or absence of lupus nephritis.
    \item \textbf{Bankrupt}~\cite{10.1093/bioinformatics/btab727}. A binary classification dataset containing 50 company records with 6 financial ratios, used to predict bankruptcy status.
\end{itemize}
\paragraph{Regression}
\begin{itemize}
    \item \textbf{Diabetes}~\cite{efron2004least}. Contains 442 samples with 10 features from medical records, used to predict a quantitative measure of disease progression one year after baseline.
    \item \textbf{Servo}~\cite{michie1995machine}. A small dataset with 167 samples and 4 input variables describing servo motor configurations, used to predict the motor’s response speed.
    \item \textbf{Friedman}~\cite{friedman1991multivariate}. A synthetic regression dataset with 500 samples and 10 features, designed for benchmarking non-linear regression methods.
\end{itemize}
\paragraph{Clustering}
\begin{itemize}
    \item \textbf{Mall}. A dataset with 200 customer records including demographic attributes and annual spending, used to segment customers into market clusters.
    \item \textbf{Wholesale}. Contains 440 observations of annual spending in different product categories for wholesale clients, used for customer segmentation.
    \item \textbf{Moon}. A synthetic two-dimensional dataset of 300 samples arranged in two interleaving half-moon shapes, commonly used to test clustering algorithms on non-linearly separable data.
\end{itemize}


To emulate realistic low-resource conditions, we refrain from applying any data augmentation or synthetic sample generation. All datasets are standardized using z-score normalization to place features on a comparable scale. The preprocessing choice also mitigates the risk of LLMs leveraging memorized raw feature distributions from prior exposure in their training data.

\subsection{Models}
We evaluate three categories of models: (i) LLMs configured via ICL, (ii) traditional ML baselines and (iii) TFMs.

\paragraph{Large Language Models (LLMs).}
The following LLMs are assessed in few-shot prompting settings: \textbf{GPT-5}, \textbf{GPT-5-Thinking}, \textbf{GPT-4o}, \textbf{GPT-o3 (OpenAI)}, \textbf{Gemini (Google DeepMind)} and \textbf{DeepSeek (DeepSeek AI)}.

LLMs are prompted with structured input-output examples using plain-text format. For classification and regression, few-shot examples are provided in tabular text format (e.g., feature names followed by their corresponding values and labels). For clustering, raw feature vectors are presented, with the model asked to identify patterns and assign cluster identifiers.

\paragraph{Machine Learning Baselines.}

We include the following ML algorithms as baselines. All models are trained using default hyperparameters.
\begin{itemize}
\item \textbf{Logistic / Liner Regression}: For classification and regression tasks, respectively.
\item \textbf{Random Forest (RF)}: An ensemble of decision trees for classification and regression.
\item \textbf{LightGBM}: A gradient-boosted decision tree model.
\item \textbf{CatBoost}: A gradient-boosted decision tree algorithm with native categorical feature handling.
\item \textbf{K-Means Clustering}: Partition-based clustering algorithm.
\item \textbf{Agglomerative Clustering}: Hierarchical clustering method.
\item \textbf{DBSCAN}: Density-based clustering algorithm.
\item \textbf{Gaussian Mixture Models (GMM)}: Probabilistic clustering approach based on Gaussian distributions.
\item \textbf{TabNet + KMeans}: TabNet~\cite{arik2021tabnet} feature extraction followed by K-Means clustering.
\end{itemize}

\paragraph{Tabular Foundational Models.}
We evaluate two recent foundational approaches designed specifically for tabular data. \textbf{TabPFN}~\cite{hollmann2022tabpfn} is a transformer-based meta-learning model trained to directly predict labels for small tabular datasets without task-specific gradient updates. It is trained on synthetic tasks drawn from a prior distribution, enabling it to perform few-shot inference on unseen datasets in a single forward pass. TabPFN can handle both classification and regression tasks. \textbf{TabICL}~\cite{qu2025tabicl} extends the in-context learning paradigm to tabular data by combining a transformer encoder with a specialized prompt construction strategy. It embeds both features and labels into a unified representation space, which enables the model to perform inference by conditioning on a set of training examples provided as context.

\subsection{Metrics}
The evaluation metrics are chosen according to the nature of each task:
\begin{itemize}
    \item \textbf{Classification}: Accuracy, Precision, Recall, and F1-Score.
    \item \textbf{Regression}: Mean Absolute Error (MAE), Mean Squared Error (MSE), Coefficient of Determination ($\text{R}^2$ Score).
    \item \textbf{Clustering}: Silhouette Coefficient~\cite{rousseeuw1987silhouettes}, Davies–Bouldin Index~\cite{davies2009cluster}, Calinski–Harabasz Index~\cite{calinski1974dendrite}.
\end{itemize}

For LLMs, outputs are post-processed through rule-based matching or semantic similarity measures to align predicted labels with ground-truth values.

\section{Experiments}
\label{sec:experimental}

\subsection{Classification Results}

\begin{table}[t!]
\centering
\caption{Classification Performance Across Datasets}
\label{tab:classification_results}
\begin{tabular}{llcccc}
\hline
\textbf{Dataset} & \textbf{Model} & \textbf{Accuracy} & \textbf{F1-Score} & \textbf{Precision} & \textbf{Recall} \\
\hline
\multirow{12}{*}{\textbf{Iris}} 
& Logistic Regression & 0.9333 & 0.9333 & 0.9333 & 0.9333 \\
& Random Forest       & 0.9333 & 0.9333 & 0.9333 & 0.9333 \\
& LightGBM            & 0.9000 & 0.8997 & 0.9024 & 0.9000 \\
& CatBoost            & 0.9333 & 0.9333 & 0.9333 & 0.9333 \\
& TabPFN              & 1.0000 & 1.0000 & 1.0000 & 1.0000 \\
& TabICL              & 0.9667 & 0.9666 & 0.9697 & 0.9667 \\
& DeepSeek            & 0.9667 & 0.9666 & 0.9697 & 0.9667 \\
& Gemini              & 0.9333 & 0.9327 & 0.9444 & 0.9333 \\
& GPT-4o              & 0.9667 & 0.9666 & 0.9697 & 0.9667 \\
& GPT-o3              & 0.9333 & 0.9327 & 0.9444 & 0.9333 \\
& GPT-5               & 0.9667 & 0.9666 & 0.9697 & 0.9667 \\
& GPT-5t              & 0.9333 & 0.9333 & 0.9333 & 0.9333 \\
\hline
\multirow{12}{*}{\textbf{Lupus}} 
& Logistic Regression & 0.8889 & 0.8903 & 0.9136 & 0.8889 \\
& Random Forest       & 0.7778 & 0.7806 & 0.8025 & 0.7778 \\
& LightGBM            & 0.8889 & 0.8903 & 0.9136 & 0.8889 \\
& CatBoost            & 0.7778 & 0.7806 & 0.8025 & 0.7778 \\
& TabPFN              & 0.8333 & 0.8349 & 0.8417 & 0.8333 \\
& TabICL              & 0.6667 & 0.6708 & 0.6914 & 0.6667 \\
& DeepSeek            & 0.7222 & 0.7249 & 0.7319 & 0.7222 \\
& Gemini              & 0.6667 & 0.6708 & 0.6914 & 0.6667 \\
& GPT-4o              & 0.8333 & 0.8349 & 0.8833 & 0.8333 \\
& GPT-o3              & 0.7222 & 0.7010 & 0.7282 & 0.7222 \\
& GPT-5               & 0.7778 & 0.7806 & 0.8025 & 0.7778 \\
& GPT-5t              & 0.8333 & 0.8305 & 0.8333 & 0.8333 \\
\hline
\multirow{12}{*}{\textbf{Bankrupt}} 
& Logistic Regression & 1.0000 & 1.0000 & 1.0000 & 1.0000 \\
& Random Forest       & 0.9000 & 0.8990 & 0.9167 & 0.9000 \\
& LightGBM            & 0.8000 & 0.7917 & 0.8571 & 0.8000 \\
& CatBoost            & 0.9000 & 0.8990 & 0.9167 & 0.9000 \\
& TabPFN              & 1.0000 & 1.0000 & 1.0000 & 1.0000 \\
& TabICL              & 1.0000 & 1.0000 & 1.0000 & 1.0000 \\
& DeepSeek            & 1.0000 & 1.0000 & 1.0000 & 1.0000 \\
& Gemini              & 0.7000 & 0.6970 & 0.7083 & 0.7000 \\
& GPT-4o              & 0.9000 & 0.8990 & 0.9167 & 0.9000 \\
& GPT-o3              & 0.8000 & 0.7917 & 0.8571 & 0.8000 \\
& GPT-5               & 1.0000 & 1.0000 & 1.0000 & 1.0000 \\
& GPT-5t              & 0.8000 & 0.8000 & 0.8000 & 0.8000 \\
\hline
\end{tabular}
\end{table}

Table~\ref{tab:classification_results} summarizes the classification performance of traditional ML models, LLM-based ICL approaches and TFMs. Across the three benchmark datasets (Iris, Lupus, Bankrupt), we observe that LLMs are capable of delivering competitive accuracy and F1-scores in several cases, often closely following the best-performing traditional baselines.

For the Iris dataset, where the class structure is well-defined and the number of samples is small, most LLMs and the foundational models (TabPFN, TabICL) achieve accuracies above 0.96, outperforming ML baselines. This suggests that in low-dimensional, clean and balanced classification problems, LLMs can act as universal approximator baselines without significant loss in performance.

In the Lupus dataset, which presents class imbalance and higher feature noise, LLM performance is more variable. Models like GPT-4o and GPT-5t match the accuracy of strong classical models (0.8333), others such as Deepseek and Gemini drop to 0.7222 and 0.6667, respectively. Nevertheless, even the foundational models in this dataset struggle to achieve high predictive accuracy, with traditional baselines outperforming recent developments.

The Bankrupt dataset presents a scenario with near-separable classes. Here, LLMs like GPT-5 and DeepSeek achieve perfect accuracy, matching TabPFN, TabICL and Logistic Regression. However, other LLMs (e.g., GPT-5t) drop to 0.8000, highlighting that while peak performance is possible, it is not guaranteed across LLM variants.

Overall, the classification results provide the strongest evidence for the viability of LLMs as universal approximator baselines. Specifically, in discrete label prediction tasks, they can reach accuracy comparable to purpose-built models with no gradient-based training. 


\subsection{Regression Results} 
\begin{table}[t!]
\centering
\caption{Regression Performance Across Datasets. Models with `--' indicate that results could not be obtained due to exceeding the maximum context length.}
\label{tab:regression_results}
\begin{tabular}{llccc}
\hline
\textbf{Dataset} & \textbf{Model} & \textbf{MSE} & \textbf{MAE} & \textbf{R\textsuperscript{2}} \\
\hline
\multirow{11}{*}{\textbf{Diabetes}} 
& Linear Regression & 2900.1936 & 42.7941 & 0.4526 \\
& Random Forest     & 2963.0138 & 44.1733 & 0.4407 \\
& LightGBM          & 3203.0861 & 44.8512 & 0.3954 \\
& CatBoost          & 2875.1092 & 43.5463 & 0.4573 \\
& TabPFN            & 2606.9805 & 40.4446 & 0.5079 \\
& DeepSeek          & 6206.5393 & 63.0337 & -0.1715 \\
& Gemini            & --        & --      & --     \\
& GPT-4o            & --        & --      & --     \\
& GPT-o3            & 12761.3951 & 93.8290 & -1.4087 \\
& GPT-5             & --        & --      & --     \\
& GPT-5T            & 6558.6966 & 61.8652 & -0.2379 \\
\hline
\multirow{11}{*}{\textbf{Servo}} 
& Linear Regression & 0.8628 & 0.7648 & 0.6433 \\
& Random Forest     & 0.2848 & 0.2825 & 0.8822 \\
& LightGBM          & 0.6964 & 0.5989 & 0.7121 \\
& CatBoost          & 0.1143 & 0.2276 & 0.9527 \\
& TabPFN            & 0.0714 & 0.1900 & 0.9705 \\
& DeepSeek          & 1.3660 & 0.5888 & 0.4353 \\
& Gemini            & --     & --     & --     \\
& GPT-4o            & 1.9826 & 0.7960 & 0.1804 \\
& GPT-o3            & 2.4303 & 0.8906 & -0.0047 \\
& GPT-5             & 2.9268 & 0.8210 & -0.2099 \\
& GPT-5T            & 0.2629 & 0.3459 & 0.8913 \\
\hline
\multirow{11}{*}{\textbf{Friedman}} 
& Linear Regression & 0.8628 & 0.7648 & 0.6433 \\
& Random Forest     & 0.2848 & 0.2825 & 0.8822 \\
& LightGBM          & 0.6964 & 0.5989 & 0.7121 \\
& CatBoost          & 0.1143 & 0.2276 & 0.9527 \\
& TabPFN            & 0.0714 & 0.1900 & 0.9705 \\
& DeepSeek          & 18.6983 & 3.5771 & 0.0760 \\
& Gemini            & --     & --     & --     \\
& GPT-4o            & 19.9051 & 3.4183 & 0.0164 \\
& GPT-o3            & 25.6578 & 3.8822 & -0.2679 \\
& GPT-5             & 18.0726 & 3.2401 & 0.1069 \\
& GPT-5T            & 16.3428 & 3.3819 & 0.1924 \\
\hline
\end{tabular}
\end{table}

The results from Regression tasks are summarized in Table~\ref{tab:regression_results}, where a significant performance gap between LLM-based approaches and established regression algorithms is evident. Specifically, high-performing classical models (e.g., CatBoost, TabPFN) achieve $R^2$ values above 0.95, while most LLMs lag substantially, with some producing negative $R^2$ scores, i.e. worse than a constant mean predictor. This indicates that LLMs in their current form do not consistently capture the continuous input–output mapping required for accurate regression.

The Diabetes dataset illustrates this limitation sharply: TabPFN reaches an $R^2$ of 0.5079 with competitive error metrics, while LLM-based outputs range from marginally positive to strongly negative $R^2$ values (e.g., GPT-o3 at –1.4087). Similar trends are observed in Servo, where only GPT-5t achieves competitive performance (0.8913 $R^2$) and most LLMs underperform basic linear regression.

An important observation is that the performance gap often widens with larger datasets or those requiring fine-grained continuous prediction. Unlike classification, where token-level categorical predictions align with the discrete nature of the task, continuous value prediction demands numerical precision that is difficult to maintain through autoregressive token generation. Furthermore, LLMs lack an explicit objective function tied to regression error, making their outputs prone to scaling errors and reduced stability when more data is present.

Overall, regression results suggest that LLMs, in their current form, are not reliable universal approximators for continuous-valued functions. While they may occasionally produce competitive numbers on small, low-noise datasets, their performance is inconsistent and generally inferior to specialised regressors, even when used only as fast baselines.

\subsection{Clustering Results}
\begin{table}[t!]
\centering
\caption{Clustering Performance Across Datasets}
\label{tab:clustering_results}
\begin{tabular}{llccc}
\hline
\textbf{Dataset} & \textbf{Method} & \textbf{Silhouette} & \textbf{Davies-Bouldin} & \textbf{Calinski-Harabasz} \\
\hline
\multirow{11}{*}{\textbf{Mall}} 
& KMeans            & 0.3348 & 1.0140 & 73.1003 \\
& Agglomerative     & 0.3102 & 1.0729 & 70.2271 \\
& DBSCAN            & 0.0120 & 1.3894 & 12.0994 \\
& GMM               & 0.3075 & 1.0474 & 64.6370 \\
& TabNet+KMeans     & 0.4078 & 0.9837 & 149.2396 \\
& DeepSeek          & 0.0846 & 3.0995 & 35.6597 \\
& Gemini            & 0.2935 & 1.5502 & 43.4959 \\
& GPT-4o            & 0.0769 & 1.2021 & 55.8180 \\
& GPT-o3            & 0.1539 & 1.4612 & 68.6584 \\
& GPT-5             & 0.2533 & 0.8914 & 52.8077 \\
& GPT-5t            & 0.4241 & 0.7532 & 143.1577 \\
\hline
\multirow{11}{*}{\textbf{Wholesale}} 
& KMeans            & 0.3494 & 1.1475 & 143.8710 \\
& Agglomerative     & 0.2670 & 0.9245 & 120.1315 \\
& DBSCAN            & 0.1958 & 1.5794 & 39.0643 \\
& GMM               & 0.1633 & 1.7059 & 74.0302 \\
& TabNet+KMeans     & 0.4911 & 0.5897 & 503.5398 \\
& DeepSeek          & -0.2791 & 4.6087 & 2.7619 \\
& Gemini            & 0.0562 & 5.1958 & 4.5278 \\
& GPT-4o            & -0.0986 & 11.5882 & 12.3377 \\
& GPT-o3            & 0.0746 & 5.1160 & 4.9525 \\
& GPT-5             & 0.0410 & 10.7528 & 4.5089 \\
& GPT-5t            & -0.0830 & 5.1255 & 6.0512 \\
\hline
\multirow{11}{*}{\textbf{Moon}} 
& KMeans            & 0.4925 & 0.8109 & 276.1942 \\
& Agglomerative     & 0.4524 & 0.8671 & 225.0146 \\
& DBSCAN            & 0.3776 & 1.0396 & 167.2785 \\
& GMM               & 0.4926 & 0.8107 & 275.9300 \\
& TabNet+KMeans     & 0.7168 & 0.4603 & 541.8967 \\
& DeepSeek          & 0.1330 & 1.3591 & 31.1146 \\
& Gemini            & 0.4178 & 0.7437 & 188.0378 \\
& GPT-4o            & 0.0591 & 7.6123 & 1.9378 \\
& GPT-o3            & 0.0363 & 3.8679 & 10.4935 \\
& GPT-5             & 0.0469 & 3.8638 & 11.1680 \\
& GPT-5t            & 0.0870 & 2.8627 & 19.8646 \\
\hline
\end{tabular}
\end{table}

Clustering performance, presented in Table~\ref{tab:clustering_results}, underscores further limitations of LLMs as unsupervised universal approximators. Standard clustering algorithms consistently outperform LLM-derived embeddings across all three datasets.

In Mall, the best LLM configuration (GPT-5t) produces a silhouette score of 0.4241, approaching the strongest TabNet+KMeans result (0.4078) and outperforming vanilla KMeans (0.3348). However, this success does not generalise: in Wholesale, LLM-based methods frequently produce negative silhouette scores and higher Davies–Bouldin scores, indicating poor separation between clusters. Similarly, in Moon, a dataset with well-defined non-linear clusters, traditional methods like TabNet+KMeans achieve 0.7168 silhouette, while LLMs indicate significant performance degradation (e.g., GPT-4o at 0.0591).

These inconsistencies suggest that while LLM embeddings may incidentally align with cluster structure in certain small, simple datasets, they do not reliably capture the geometry of data manifolds in a way that supports unsupervised separation. This is likely due to the fact that LLM representations are trained for general semantic similarity rather than the specific variance–covariance structures exploited by clustering algorithms.

In conclusion, LLM-based methods show limited promise as universal approximators in clustering. While they can serve as exploratory baselines in low-complexity scenarios, their instability across datasets mean they cannot yet replace domain-tuned clustering pipelines for unsupervised learning.

\subsection{Ablation Study}
To better understand the factors influencing LLM performance on small tabular classification tasks, we conduct a series of controlled ablation experiments using a single stratified train/test split (70/30) for each dataset (Iris, Lupus, Bankrupt). All features are z-score normalized using training-set statistics. We compare the two strongest LLMs from our main results (GPT-5, DeepSeek) against two foundation models for tabular data (TabPFN, TabICL). 



\begin{figure}[t!]
    \centering
    \includegraphics[width=0.85\linewidth]{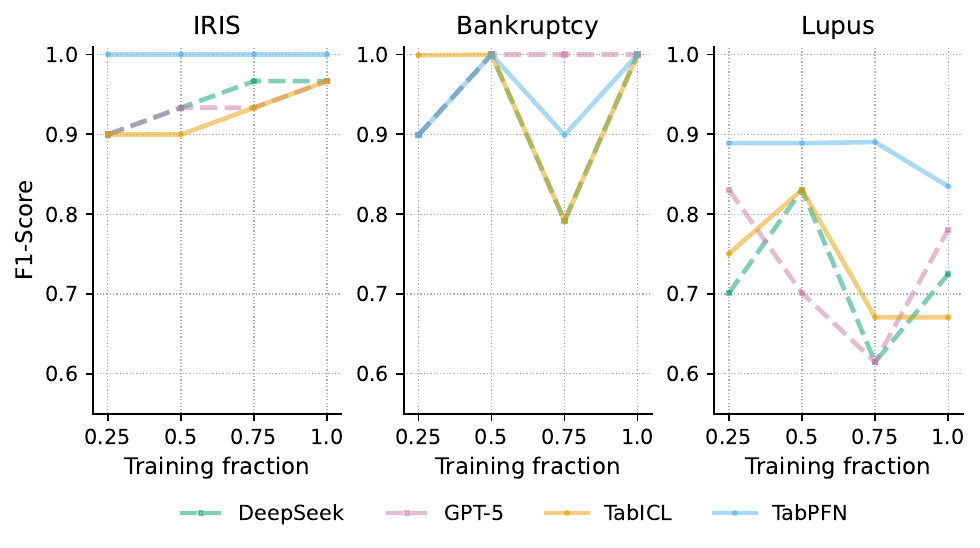}
    \caption{\textbf{F1-Score trends across training data fractions.} 
    Each panel corresponds to a dataset (IRIS, Bankruptcy, Lupus), with four models evaluated (TabICL, TabPFN, DeepSeek, GPT-5). Solid lines indicate tabular models; dashed lines indicate LLMs.}
    \label{fig:ablation-number-samples}
\end{figure}

\paragraph{Effect of Training Data Fraction.}
We vary the fraction of available training data to determine the minimum amount required for LLMs to match the performance of TFMs. Fractions of 25\%, 50\%, 75\%, and 100\% of the training split are evaluated. For TabPFN and TabICL, models are trained directly on each subsample. For LLMs, the available fraction defines the in-context example pool. Figure~\ref{fig:ablation-number-samples} presents the result of this ablation.

On IRIS, TabICL and both LLMs (DeepSeek, GPT-5) begin with the same F1 value at 0.25 ($\approx$0.8997) and ended with the same F1 at full data ($\approx$0.9667). In all IRIS cases, increasing the number of training samples led to monotonic F1 improvements, with TabPFN achieving 1.0 across all fractions, showing complete stability and no dependence on sample size.

In Bankruptcy, DeepSeek and TabPFN started with identical F1 at 0.25 ($\approx$0.8990), while GPT-5 achieved 1.0 at all fractions. From the 0.5 fraction onwards, TabICL and DeepSeek produced exactly the same F1 values, including perfect scores at 0.5 and 1.0. Here, however, more training data did not always improve performance. Both TabICL and DeepSeek dropped from perfect scores at 0.5 to $\approx$0.80 at 0.75 fraction. This performance degradation likely reflects overfitting to some patterns when additional but noisy samples are introduced.

For Lupus, the differences between models are clearer than the rest datasets. TabPFN consistently outperformed others ($\approx$0.83–0.89) across all fractions, while TabICL, DeepSeek, and GPT-5 showed variability. Specifically, at 0.75 fraction, their F1 values dropped significantly ($\approx$0.61–0.67), with only partial recovery at full data (GPT-5 $\approx$0.78, DeepSeek $\approx$0.72, TabICL $\approx$0.67). As in Bankruptcy, more samples sometimes reduced F1, possibly due to increased heterogeneity.

Overall, TFMs (TabICL, TabPFN) showed more stability across fractions, with TabPFN being largely unaffected by data size, while LLMs (DeepSeek, GPT-5) exhibited dataset-specific trends.

The results of LLMs, in our experiments, show that high performance on tabular tasks can be achieved without fine-tuning, even when using relatively few examples. However, the effect of sample size is not uniform across datasets. 


\paragraph{Serialization Format}
We investigate whether the formatting of tabular data within the prompt affects LLM performance. Four serialization formats are compared: (i) CSV (default), (ii) key:value pairs, (iii) JSON Lines, and (iv) Markdown tables. This experiment is conducted on the Lupus dataset. 

\begin{table}[t!]
\centering
\caption{Performance of LLMs with different Serialization Formats on Lupus Dataset}
\label{tab:serialization_results}
\begin{tabular}{llcccc}
\hline
Model                     & Format    & Accuracy & Precision & Recall & F1-Score     \\ \hline
\multirow{4}{*}{GPT-5}    & CSV       & 0.7778   & 0.6667    & 0.8571 & 0.750  \\
                          & KEY:VALUE & 0.6667   & 0.5556    & 0.7143 & 0.625  \\
                          & JSON      & 0.7778   & 0.6667    & 0.8571 & 0.750  \\
                          & MARKDOWN  & 0.7778   & 0.6667    & 0.8571 & 0.750  \\ \hline
\multirow{4}{*}{Deepseek} & CSV       & 0.7222   & 0.625     & 0.7143 & 0.6667 \\
                          & KEY:VALUE & 0.7222   & 0.625     & 0.7143 & 0.6667 \\
                          & JSON      & 0.8333   & 0.700     & 1.0000 & 0.8235 \\
                          & MARKDOWN  & 0.7222   & 0.625     & 0.7143 & 0.6667 \\ \hline
\end{tabular}
\end{table}

The results indicate that the choice of serialization format can influence LLM performance, though the extent of this effect appears model-dependent. For GPT-5, performance remains largely stable across CSV, JSON, and Markdown, suggesting that the model is relatively robust to variations in tabular data formatting. In contrast, Deepseek demonstrates a clearer sensitivity to format, with JSON yielding noticeably better results compared to other formats. This suggests that for certain models, providing data in structured JSON form could enhance predictive performance.

\section{Conclusion}
\label{sec:conclusions}
This paper explores the use of LLMs as function approximators on small tabular datasets across classification, regression and clustering tasks, using ICL without explicit fine-tuning. Our results suggest that LLMs can offer competitive performance in classification tasks, in some cases approaching the accuracy of traditional ML models and tabular foundation models, demonstrating their potential as zero-training baselines for discrete prediction problems in settings where rapid prototyping or minimal setup is desired.

For regression tasks, however, LLM performance was generally less reliable. The models produced lower accuracy compared to standard regressors. This may be due to the difficulty of producing accurate continuous outputs via autoregressive generation. In clustering, the results were similarly mixed, which we attribute to no actual ICL in this setting.

The ablation study showed that LLM behavior can vary depending on dataset characteristics and the number of in-context examples. In some cases, more data did not consistently lead to improved results, pointing to potential sensitivity to data distribution or input representation.

Overall, although LLMs may not currently serve as universal approximators across all learning tasks, they can be useful tools for classification over tabular data when ease of use and fast iteration are important. Further work is needed to better understand their limitations in regression and clustering and to explore ways to extend LLMs for more reliable performance on structured data tasks.
%
%
%
\bibliographystyle{splncs04}
\bibliography{main}

\begin{thebibliography}{10}
\providecommand{\url}[1]{\texttt{#1}}
\providecommand{\urlprefix}{URL }
\providecommand{\doi}[1]{https://doi.org/#1}

\bibitem{achiam2023gpt}
Achiam, J., Adler, S., Agarwal, S., Ahmad, L., Akkaya, I., Aleman, F.L., Almeida, D., Altenschmidt, J., Altman, S., Anadkat, S., et~al.: Gpt-4 technical report. arXiv preprint arXiv:2303.08774  (2023)

\bibitem{anghel2018benchmarking}
Anghel, A., Papandreou, N., Parnell, T., De~Palma, A., Pozidis, H.: Benchmarking and optimization of gradient boosting decision tree algorithms. arXiv preprint arXiv:1809.04559  (2018)

\bibitem{arik2021tabnet}
Arik, S.{\"O}., Pfister, T.: Tabnet: Attentive interpretable tabular learning. In: Proceedings of the AAAI conference. vol.~35, pp. 6679--6687 (2021)

\bibitem{briola2024federated}
Briola, E., Nikolaidis, C.C., Perifanis, V., Pavlidis, N., Efraimidis, P.: A federated explainable ai model for breast cancer classification. In: Proceedings of the 2024 European Interdisciplinary Cybersecurity Conference. pp. 194--201 (2024)

\bibitem{brown2020language}
Brown, T., Mann, B., Ryder, N., Subbiah, M., Kaplan, J.D., Dhariwal, P., Neelakantan, A., Shyam, P., Sastry, G., Askell, A., et~al.: Language models are few-shot learners. Advances in neural information processing systems  \textbf{33},  1877--1901 (2020)

\bibitem{calinski1974dendrite}
Cali{\'n}ski, T., Harabasz, J.: A dendrite method for cluster analysis. Communications in Statistics-theory and Methods  \textbf{3}(1),  1--27 (1974)

\bibitem{cheng2025realistic}
Cheng, Z.J., Jia, Z.Y., Zhou, Z., Li, Y.F., Guo, L.Z.: Realistic evaluation of tabpfn v2 in open environments. arXiv preprint arXiv:2505.16226  (2025)

\bibitem{davies2009cluster}
Davies, D.L., Bouldin, D.W.: A cluster separation measure. IEEE transactions on pattern analysis and machine intelligence (2),  224--227 (2009)

\bibitem{dong2024large}
Dong, H., Wang, Z.: Large language models for tabular data: Progresses and future directions. In: Proceedings of the 47th International ACM SIGIR Conference on Research and Development in Information Retrieval. pp. 2997--3000 (2024)

\bibitem{dong-etal-2024-survey}
Dong, Q., Li, L., Dai, D., ...~Sui, Z.: A survey on in-context learning. In: Proceedings of the 2024 Conference on Empirical Methods in NLP. pp. 1107--1128. ACL, Miami, Florida, USA (Nov 2024). \doi{10.18653/v1/2024.emnlp-main.64}

\bibitem{efron2004least}
Efron, B., Hastie, T., Johnstone, I., Tibshirani, R.: Least angle regression  (2004)

\bibitem{fisher1936use}
Fisher, R.A.: The use of multiple measurements in taxonomic problems. Annals of eugenics  \textbf{7}(2),  179--188 (1936)

\bibitem{friedman1991multivariate}
Friedman, J.H.: Multivariate adaptive regression splines. The annals of statistics  \textbf{19}(1),  1--67 (1991)

\bibitem{gorishniy2021revisiting}
Gorishniy, Y., Rubachev, I., Khrulkov, V., Babenko, A.: Revisiting deep learning models for tabular data. Advances in neural information processing systems  \textbf{34},  18932--18943 (2021)

\bibitem{he2021automl}
He, X., Zhao, K., Chu, X.: Automl: A survey of the state-of-the-art. Knowledge-based systems  \textbf{212},  106622 (2021)

\bibitem{hollmann2022tabpfn}
Hollmann, N., M{\"u}ller, S., Eggensperger, K., Hutter, F.: Tabpfn: A transformer that solves small tabular classification problems in a second. arXiv:2207.01848  (2022)

\bibitem{hornik1991approximation}
Hornik, K.: Approximation capabilities of multilayer feedforward networks. Neural networks  \textbf{4}(2),  251--257 (1991)

\bibitem{jaitly2023towards}
Jaitly, S., Shah, T., ...~Grewal, R.S.: Towards better serialization of tabular data for few-shot classification with large language models. arXiv:2312.12464  (2023)

\bibitem{kratsios2021universal}
Kratsios, A., Zamanlooy, B., Liu, T., Dokmani{\'c}, I.: Universal approximation under constraints is possible with transformers. arXiv preprint arXiv:2110.03303  (2021)

\bibitem{liu2024deepseek}
Liu, A., Feng, B., Xue, B., Wang, B., Wu, B., Lu, C., Zhao, C., Deng, C., Zhang, C., Ruan, C., et~al.: Deepseek-v3 technical report. arXiv preprint:2412.19437  (2024)

\bibitem{lu2025fine}
Lu, P., Zhang, P.., Liu, T.: Fine-tuning pre-trained large language models for price prediction on network freight platforms. Mathematics  \textbf{13}(15), ~2504 (2025)

\bibitem{lu2025large}
Lu, W., Zhang, J., Fan, J., Fu, Z., Chen, Y., Du, X.: Large language model for table processing: A survey. Frontiers of Computer Science  \textbf{19}(2),  192350 (2025)

\bibitem{michie1995machine}
Michie, D., Spiegelhalter, D.J., Taylor, C.C., Campbell, J.: Machine learning, neural and statistical classification. Ellis Horwood (1995)

\bibitem{qu2025tabicl}
Qu, J., Holzm{\~A}{\v{z}}ller, D., Varoquaux, G., Morvan, M.L.: Tabicl: A tabular foundation model for in-context learning on large data. arXiv preprint:2502.05564  (2025)

\bibitem{10.1093/bioinformatics/btab727}
Romano, J.D., Le, T.T., La~Cava, W., Gregg, J.T., Goldberg, D.J., Chakraborty, P., Ray, N.L., Himmelstein, D., Fu, W., Moore, J.H.: Pmlb v1.0: an open-source dataset collection for benchmarking machine learning methods. Bioinformatics  \textbf{38}(3),  878--880 (10 2021). \doi{10.1093/bioinformatics/btab727}

\bibitem{rousseeuw1987silhouettes}
Rousseeuw, P.J.: Silhouettes: a graphical aid to the interpretation and validation of cluster analysis. Jour. of computational and applied mathematics  \textbf{20},  53--65 (1987)

\bibitem{shamshiri2024context}
Shamshiri, A., Ryu, K.R., Park, J.Y.: In-context learning for long-context sentiment analysis on infrastructure project opinions. arXiv preprint:2410.11265  (2024)

\bibitem{shin2022effect}
Shin, S., Lee, S.W., Ahn, H.., Ha, J.W., et~al.: On the effect of pretraining corpora on in-context learning by a large-scale language model. arXiv:2204.13509  (2022)

\bibitem{team2023gemini}
Team, G., Anil, R., Borgeaud, S., Alayrac, J.B., Yu, J., Soricut, R., Schalkwyk, J., Dai, A.M., Hauth, A., Millican, K., et~al.: Gemini: a family of highly capable multimodal models. arXiv preprint arXiv:2312.11805  (2023)

\bibitem{vaswani2017attention}
Vaswani, A., Shazeer, N., Parmar, N., Uszkoreit, J., Jones, L., Gomez, A.N., Kaiser, {\L}., Polosukhin, I.: Attention is all you need. Advances in neural information processing systems  \textbf{30} (2017)

\bibitem{wen2024innovative}
Wen, Z., Zhan, J., Zhao, B., Luo, R.: Innovative application of deep learning and large model. In: 2024 5th International Conference on Information Science, Parallel and Distributed Systems (ISPDS). pp. 648--654. IEEE (2024)

\end{thebibliography}

\end{document}